\documentclass[runningheads]{llncs}

\usepackage{eccv}

\usepackage{graphicx}
\usepackage{booktabs}
\usepackage{multirow}
\usepackage{array}
\usepackage[table]{xcolor}
\usepackage{wrapfig}
\usepackage{tikz}
\usepackage{comment,verbatim}
\usepackage{amsmath,amssymb}
\usepackage{subcaption}
\usepackage{pifont}
\usepackage{algorithm}
\usepackage{algorithmic}
\usepackage{color}

\usepackage[accsupp]{axessibility}

\usepackage{hyperref}
\usepackage[capitalize]{cleveref}

\usepackage{orcidlink}

\newcommand{\cmark}{\ding{51}}
\DeclareMathOperator*{\argmin}{arg\,min}

\begin{document}

\title{\texorpdfstring{HieDG: A \underline{Hie}rarchical \underline{D}iscrete \underline{G}eometry-Guided Framework for Multi-Animal Tracking}{HieDG: A Hierarchical Discrete Geometry-Guided Framework for Multi-Animal Tracking}}

\titlerunning{HieDG for Multi-Animal Tracking}

\author{Chenxun Deng\inst{1,2}\orcidlink{0009-0007-8423-9172} \and
Zhongde Zhang\inst{3}\orcidlink{0009-0001-0914-6087} \and
Ye Yuan\inst{1,4}\orcidlink{0009-0003-6767-5854} \and
Chengyang Zhang\inst{5}\orcidlink{0000-0003-3658-5779} \and
Yifan Zhang\inst{6}\orcidlink{0000-0002-9190-3509} \and
Bohao Chen\inst{1,4}\orcidlink{0009-0002-4628-606X} \and
Hongying Yan\inst{7}\orcidlink{0009-0008-0773-3131} \and
Hang Zhou\inst{8}\orcidlink{0000-0001-8662-9376} \and
Hua Han\inst{1,9}\orcidlink{0000-0003-4713-4631} \and
Xi Chen\inst{1}$^{*}$\orcidlink{0000-0002-6922-2838}}

\authorrunning{C. Deng et al.}

\institute{
State Key Laboratory of Brain Cognition and Brain-inspired Intelligence Technology,
Institute of Automation, Chinese Academy of Sciences
\and
School of Artificial Intelligence,
University of Chinese Academy of Sciences
\and
School of Technology,
Beijing Forestry University
\and
School of Advanced Interdisciplinary Sciences,
University of Chinese Academy of Sciences
\and
College of Computer Science,
Sichuan University
\and
C²DL, Institute of Automation,
Chinese Academy of Sciences.
\and
School of Automation,
Chongqing University
\and
Faculty of Life and Health Sciences,
Shenzhen University of Advanced Technology
\and
School of Future Technology,
University of Chinese Academy of Sciences
\\
\email{xi.chen@ia.ac.cn}
}

\maketitle

\vspace{-8mm}

\begin{abstract}
Multi-animal tracking (MAT) is critical for wildlife monitoring and behavioral analysis, yet remains challenging due to uniform appearance, high density, and irregular motion. Existing methods typically follow heuristic- or query-based paradigms: the former relies on handcrafted geometric associations without end-to-end optimization, whereas the latter enables joint optimization but relies heavily on appearance embeddings. In such conditions, continuous geometric embeddings can be unstable, as small coordinate perturbations may disproportionately alter cross-frame attention weights, degrading identity association performance. To address this limitation, we propose \textbf{HieDG}, a \textbf{Hie}rarchical \textbf{D}iscrete \textbf{G}eometry-guided tracking framework that reformulates geometric dynamics as structured discrete representations within a query-based tracker. Instead of directly using raw geometric signals, HieDG employs a two-stage residual codebook to discretize position, scale, and velocity cues, transforming unstable continuous geometry into structured, stable discrete tokens. These tokens are aligned with visual embeddings and integrated into the tracking queries to enhance identity consistency. Extensive experiments on animal-specific benchmarks (AnimalTrack, BFT, and BuckTales) demonstrate state-of-the-art association performance with significant improvements in HOTA, AssA, and IDF1. Additional evaluations on generic multi-object tracking benchmarks, including DanceTrack and SportsMOT, show competitive performance, indicating the broader applicability of discretized geometric modeling beyond animal-specific scenarios.

\end{abstract}

\section{Introduction}
\label{sec:intro}

\vspace{-4mm}
Multi-animal tracking (MAT), a representative yet challenging multi-object tracking problem, plays a pivotal role in biology, ecology, and wildlife research, supporting applications such as animal monitoring~\cite{jetz2022biological,liu2020common}, population estimation~\cite{bertorelle2022genetic}, and behavior analysis~\cite{pereira2022sleap}. 
MAT involves accurately locating all animals within a video stream while consistently maintaining their identities throughout the sequence. 
Despite remarkable advances in human and vehicle tracking~\cite{LIU2024109161,zhang2023animaltrack,DENG2025106794,DENG2026113669}, MAT remains technically demanding, particularly under \emph{low visual discriminability regimes}, where appearance cues provide limited identity separation and geometric dynamics exhibit stronger variability.

Driven by advances in object detection, the bottleneck in multi-object tracking has primarily shifted from accurate localization to robust and consistent identity association~\cite{Han_2024_CVPR,NEURIPS2024_c34ddd05}. To address this, two major paradigms have emerged: heuristic-based association and Transformer-driven query-based tracking. Traditional heuristic-based paradigms typically adopt a detector–tracker pipeline, relying on meticulously crafted rules (e.g., Kalman filter~\cite{welch1995introduction} and Hungarian algorithm~\cite{mills2007dynamic}) to model geometric cues for association~\cite{Shim_2025_CVPR,ByteTrack}. Despite notable progress, these methods rely on fixed association rules that are inherently incompatible with gradient-based end-to-end optimization. Consequently, they struggle to adapt to evolving visual and geometric cues, leading to suboptimal performance in complex tracking scenarios~\cite{yan2025comot}. In contrast, query-based methods formulate tracking as query propagation across frames, enabling end-to-end identity modeling via cross-frame attention~\cite{Meinhardt_2022_CVPR,zeng2022motr,motrv2,Gao_2025_CVPR}. Although effective on human and vehicle benchmarks, these methods are heavily appearance-driven. In scenarios with weak visual discriminability, such reliance renders identity association inherently fragile.

\begin{figure}[htbp]
  \centering
  \begin{subfigure}{\linewidth}
    \centering
    \includegraphics[width=0.8\linewidth]{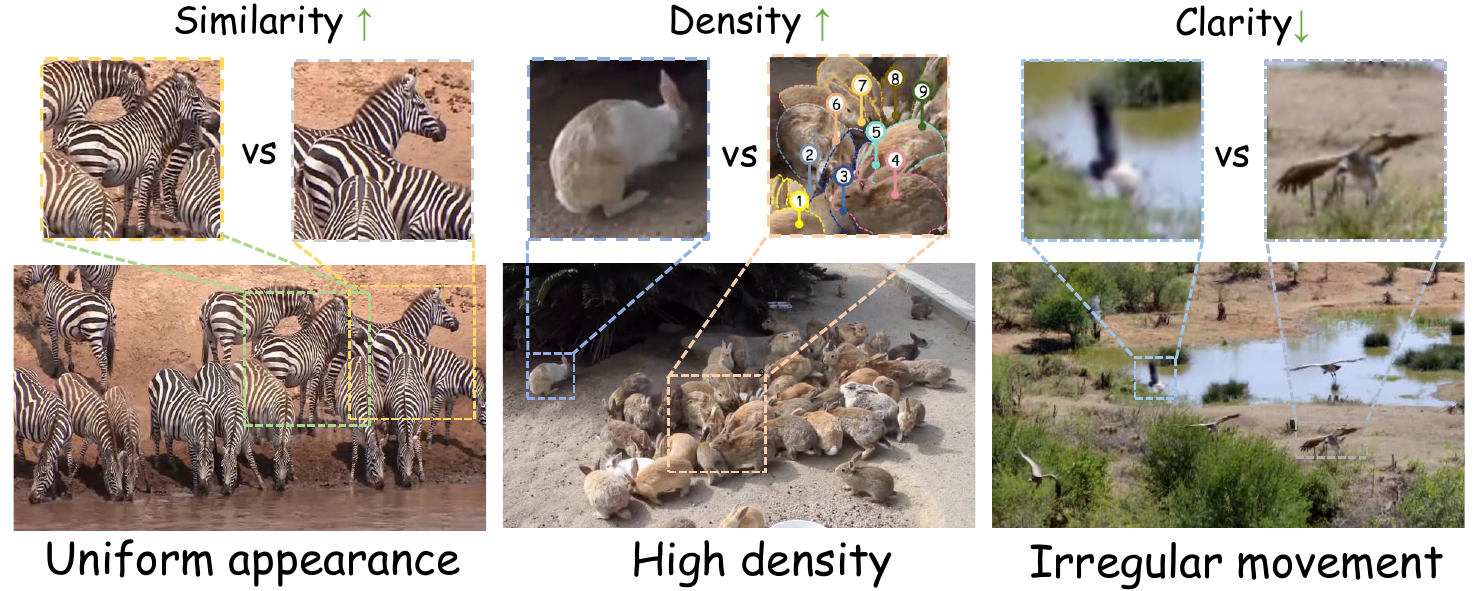}
    \caption{Challenges of animal tracking}
    \label{fig:fig1_a}
  \end{subfigure}

  \begin{subfigure}{\linewidth}
    \centering
    \includegraphics[width=0.9\linewidth]{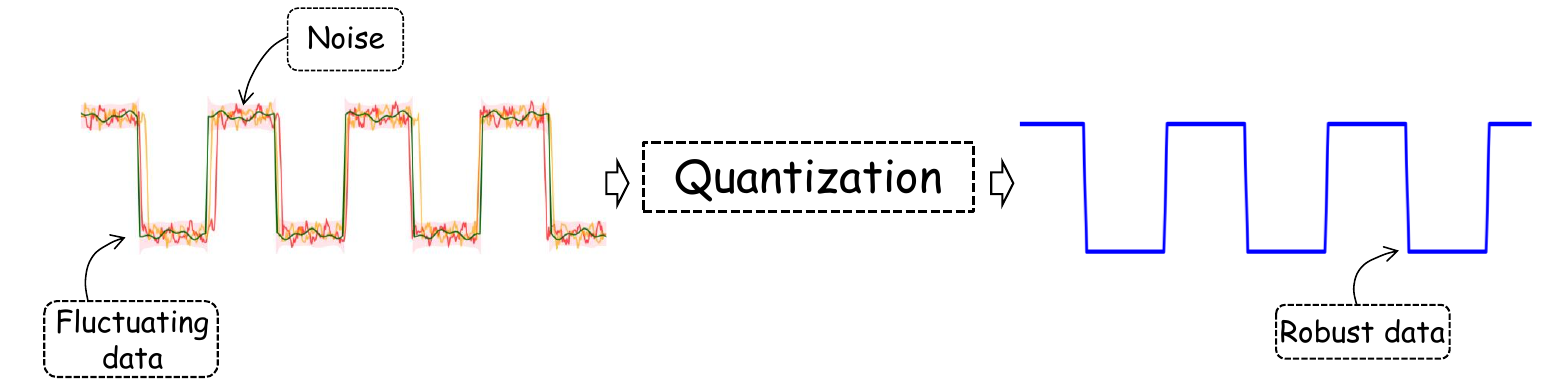}
    \caption{Illustration of data quantization}
    \label{fig:fig1_b}
  \end{subfigure}

  \caption{
  (a) Typical conditions in multi-animal tracking, including uniform appearance, high density, and irregular motion, which reduce appearance discriminability.
  (b) Quantization transforms fluctuating continuous signals into stable discrete states, suppressing perturbation-induced variability.
  }
  \label{fig:fig1}
\end{figure}

Multi-animal tracking presents precisely such a regime (Fig.~\ref{fig:fig1_a}). 
First, animals of the same species often exhibit a highly uniform appearance due to social grouping behavior, resulting in minimal inter-instance visual distinction. 
Second, animals frequently form high-density groups, where close spatial proximity and frequent overlap significantly increase identity ambiguity. 
Third, irregular and non-planar movement patterns, especially in aerial or aquatic environments, introduce strong geometric variability across frames. 
Overall, these factors place MAT in a low-discriminability, high-variability regime, in which both appearance and continuous geometric cues become less reliable for stable identity modeling.

This motivates incorporating geometric cues into end-to-end query-based tracking. However, geometric modeling under attention-based architectures is non-trivial. As observed in~\cite{zhang2023animaltrack}, camera motion, object jitter, and environmental noise introduce subtle yet persistent fluctuations in geometric signals. Although small in raw analog space, such perturbations may induce disproportionate shifts in cross-frame attention weight distributions, leading to unstable identity matching. Empirically, as we further demonstrate in Sec.~\ref{sec:comparisons}, directly incorporating continuous geometric representations can even degrade association performance relative to appearance-only baselines. This phenomenon suggests that continuous geometric embeddings may destabilize attention-based identity modeling under noisy and appearance-ambiguous conditions.

Inspired by quantization principles in signal processing—where noisy analog signals are mapped to stable discrete states (Fig.~\ref{fig:fig1_b})—we revisit geometric modeling from a representation perspective. Rather than treating geometry purely as a continuous regression signal, we introduce a discretized state-space representation for geometric dynamics. By mapping fluctuating geometric cues to a finite set of learned codewords, we constrain perturbation-induced embedding variance while preserving structural information essential for stable identity association. This discrete formulation mitigates perturbation amplification and yields more stable attention responses.


Based on this insight, we propose \textbf{HieDG}, a \textbf{Hie}rarchical \textbf{D}iscrete \textbf{G}eometry-guided framework for multi-animal tracking. HieDG employs a coarse-to-fine residual quantization strategy to discretely encode position, size, and velocity cues through independent two-level codebooks. This hierarchical design captures both coarse geometric structure and fine-grained variations, transforming unstable continuous signals into robust discrete embeddings aligned with the visual feature space. The resulting geometric tokens are fused with tracking queries, enhancing identity consistency under challenging conditions.
Our contributions are summarized as follows:

\begin{itemize}
\item We propose HieDG, a geometry-guided multi-animal tracking framework that unifies heuristic geometric reasoning with end-to-end query-based identity modeling via discrete geometric representations.

\item We design a hierarchical residual quantization strategy that encodes position, scale, and velocity cues into structured discrete tokens, reformulating geometric modeling as a robust representation-learning problem within tracking.

\item Extensive experiments on AnimalTrack, BFT, and BuckTales demonstrate state-of-the-art performance with significant improvements in HOTA, AssA, and IDF1. Additional evaluations on generic MOT benchmarks show competitive performance, indicating the general applicability of discretized geometric modeling beyond animal-specific scenarios.
\end{itemize}

\section{Related Work}
\label{sec:related_work}
\textbf{Tracking-by-Detection} has become the dominant paradigm in modern MOT, where object detection and data association are decoupled. With the rapid progress in object detection, recent efforts have increasingly focused on the association stage, aiming to construct coherent trajectories across frames. Many prior works adopt heuristic post-processing strategies to associate current detections with historical trajectories~\cite{Shim_2025_CVPR}. Early approaches, such as SORT~\cite{7533003}, rely on geometry by computing an IoU matrix between frame-wise detections and applying the Hungarian algorithm for optimal matching. ByteTrack~\cite{ByteTrack} further incorporates low-score detections to recover temporarily occluded targets with dropped confidence, significantly improving identity preservation in occlusion-heavy scenes. OC-SORT~\cite{Cao_2023_CVPR} emphasizes motion continuity by modeling trajectory consistency and smoothing motion direction, showing notable advantages in long-term occlusions and crossing scenarios. While heuristic-based paradigms perform well under certain conditions through carefully crafted geometric modeling, their fixed association rules often require manual tuning and struggle to generalize across diverse or complex scenarios. Nevertheless, their explicit use of geometric consistency provides a useful inductive bias for identity association, motivating us to integrate geometry into end-to-end trainable frameworks.

\noindent\textbf{Tracking-by-Query} has emerged as a powerful paradigm in modern MOT due to its end-to-end trainability and its capability for global optimization. By propagating object queries across frames and performing cross-frame attention, these methods unify detection and association, enabling joint learning of localization and identity assignment via gradient descent. TransTrack~\cite{sun2020transtrack} first introduces Transformers into MOT by propagating tracking queries through temporal attention. TrackFormer~\cite{Meinhardt_2022_CVPR} further unifies detection and association within a shared decoder to improve alignment. MOTR~\cite{zeng2022motr,motrv2} models identity propagation by dynamically reusing historical queries, allowing flexible handling of object births and terminations. In contrast, MOTIP~\cite{Gao_2025_CVPR} formulates MOT as an identity prediction problem trained via classification loss, avoiding explicit matching operations.

Although these approaches achieve strong performance on standard human and vehicle benchmarks, their identity modeling remains largely appearance-driven. Under conditions commonly observed in multi-animal tracking, such as uniform appearance, high density, and irregular motion, appearance embeddings alone may provide limited separability, making association more challenging. Inspired by the geometric reasoning in classical heuristic pipelines, we incorporate explicit geometric cues (e.g., position, scale, and velocity) into the query-based framework to enhance robustness under visually ambiguous and dynamically complex scenarios.

\subsection{Motivation and Overview}
\label{sec:overview}

\begin{figure*}[t]
    \centering
    \includegraphics[width=\linewidth]{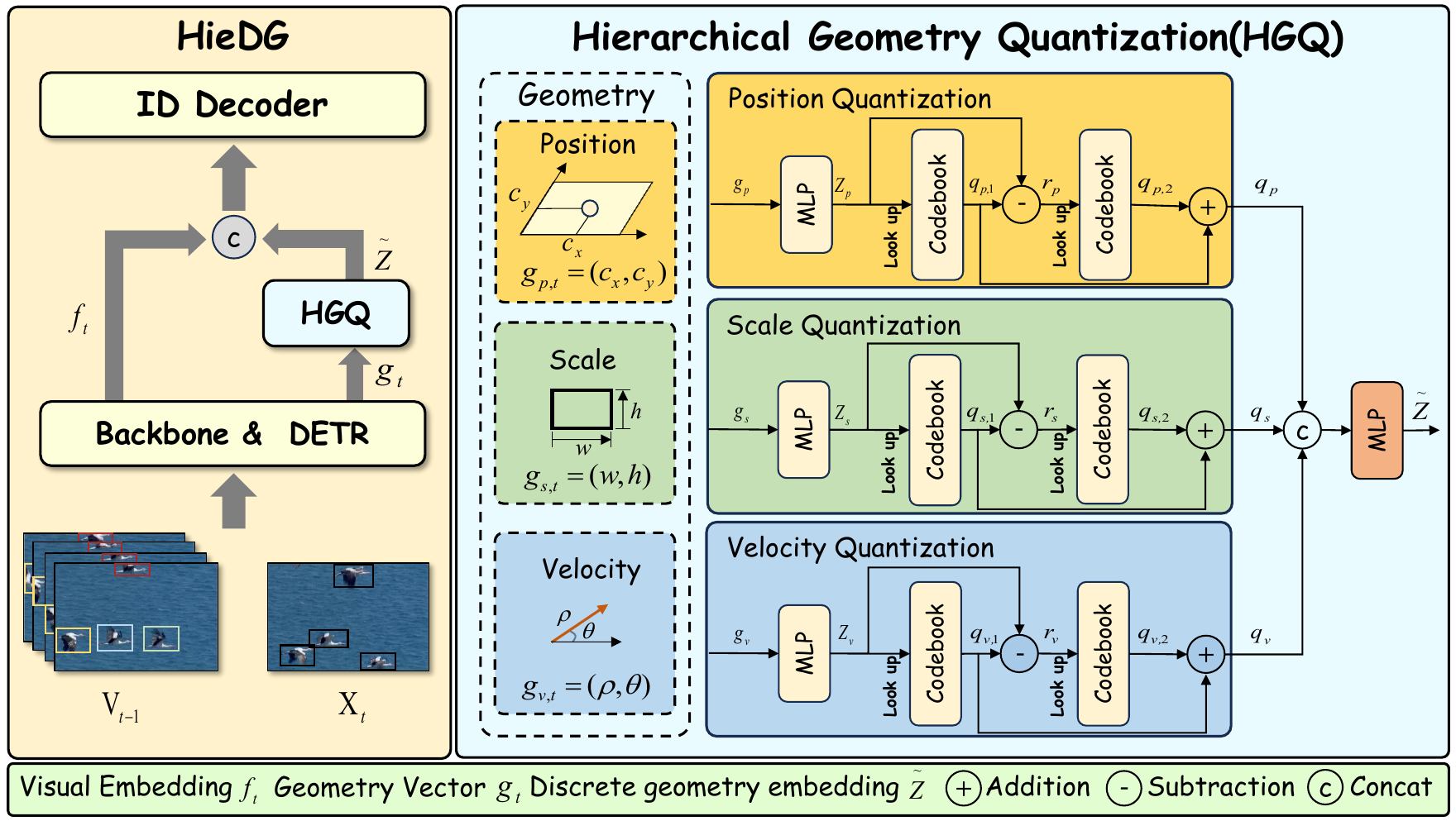}
    \caption{
    Overview of the HieDG framework. Appearance and geometric features are extracted via Deformable DETR~\cite{zhu2021deformable}. 
    The geometric vectors are first projected through MLPs and independently quantized by two-stage residual codebooks for position, size, and velocity. 
    The resulting discrete geometric embeddings are concatenated and aligned to match the visual space. 
    The combined representation is finally passed to the ID decoder for trajectory association.
    }
    \label{fig:overview}
\end{figure*}

\section{Method}

In this section, we introduce our proposed MAT framework, HieDG, which integrates robust discrete geometric cues into the query-based paradigm. We begin with the motivation and overview in Sec.~\ref{sec:overview}. Then, we detail the geometric vector construction and the hierarchical geometry quantization in Sec.~\ref{sec:gvc} and Sec.~\ref{sec:hgq}, respectively. The geometry-aware animal tracking is described in Sec.~\ref{sec:gat}. Finally, we outline the training and inference procedures in Sec.~\ref{sec:train} and Sec.~\ref{sec:inference}.

\textbf{Motivation.} In multi-animal tracking scenarios, uniform appearance, high density, and irregular motion patterns pose significant challenges to both heuristic-based and query-based tracking paradigms. Heuristic approaches, while effective at geometric modeling, lack the capacity for end-to-end global optimization, limiting their adaptability across diverse animal scenes. In contrast, query-based methods benefit from joint optimization but heavily rely on visual features, which are often non-discriminative in animal settings. This motivates a natural and intuitive question: can we combine the strengths of both paradigms by leveraging the global optimization capability of query-based frameworks to model geometric cues directly from video sequences, thereby enhancing identity association?

However, within query-based tracking frameworks, association is typically performed via cross-attention between encoded features. In real-world animal monitoring videos, camera shake, object jitter, and visual noise introduce fluctuations in geometric signals. Although subtle at the raw signal level, these variations can be amplified during embedding, leading the attention mechanism to misinterpret them as genuine geometric discrepancies—ultimately compromising association reliability. In signal processing and circuit analysis, quantization is often employed to convert perturbed analog signals into discrete and deterministic states, enhancing both stability and interpretability. Inspired by this, we seek to improve the discriminability of fluctuating geometric embeddings by discretizing them into structured discrete codeword representations. 



\textbf{Overview.}
The overall architecture of HieDG is illustrated in Fig.~\ref{fig:overview}. 
We extract object-level geometric cues (position, scale, and velocity) from Deformable DETR~\cite{zhu2021deformable} outputs and project them into latent vectors. 
Each geometric cue is independently quantized using a two-stage residual codebook in a coarse-to-fine manner, producing structured discrete tokens. 
These tokens constrain perturbation-induced embedding variance while preserving essential geometric information. 
The resulting discrete geometric embeddings are aligned with the visual feature space and fused with appearance features within the query-based tracking framework to guide robust identity association.

\subsection{Geometric Vector Construction}
\label{sec:gvc}

We adopt Deformable DETR~\cite{zhu2021deformable} as the end-to-end object detector to process each video sequence \( V=\{\mathbf{X}_t\}_{t=1}^{T} \), where \( \mathbf{X}_t \in \mathbb{R}^{H \times W \times 3} \) denotes the RGB image at frame \( t \). A CNN backbone~\cite{he2016deep} and a Transformer encoder~\cite{NIPS2017_3f5ee243} are employed to extract and enhance image features \( f_t \in \mathbb{R}^{h \times w \times c} \). Subsequently, the DETR decoder transforms detection queries into object-level predictions, including bounding box $bbox = (c_{x,t}^i,c_{y,t}^i,w_t^i,h_t^i)$ and class index. Deformable DETR directly provides decoded embeddings as object-level features $f_t^i$, eliminating the need for post-processing steps such as RoI cropping or feature alignment, obtaining dimension consistent representations across frames.

To accommodate video sequences with varying resolutions and aspect ratios, we normalize the bounding box parameters as $(c_{x,t}^i/W,c_{y,t}^i/H,w_t^i/W,h_t^i/H)$, where \( W \) and \( H \) denote the frame width and height, respectively. During training, the velocity of object \( i \) is computed by differencing the normalized center coordinates across consecutive frames, yielding Cartesian velocity components \( (v_{x,t}^i, v_{y,t}^i) \). As polar coordinates explicitly decouple two identity-relevant factors: magnitude and direction \cite{fiquet2023a}, we convert the velocity representation from Cartesian $(v_{x,t}^i,v_{y,t}^i)$ to polar coordinates\((\rho_t^i,\theta_t^i)\). This formulation decouples motion magnitude and direction, reducing entanglement between velocity components and facilitating more stable representation. As a result, the entire geometric state vector $g_t^i = (g_{p,t}^i, g_{s,t}^i, g_{v,t}^i)$ for object \( i \) at frame \( t \) is thus defined as:
\begin{equation}
g_t^i = \left( \frac{c_{x,t}^i}{W}, \frac{c_{y,t}^i}{H}, \frac{w_t^i}{W}, \frac{h_t^i}{H}, \rho_t^i, \theta_t^i \right),
\end{equation}
where \( g_t^i \) represents the geometric vector of object \( i \) at frame \( t \), including its position, size, and velocity.

\subsection{Hierarchical Geometry Quantization}
\label{sec:hgq}
Given that the constructed geometry vectors exhibit heterogeneous scales and units across dimensions, we perform residual quantization separately on position $g_{p,t}^i$, scale $g_{s,t}^i$, and velocity $g_{v,t}^i$. Each component is projected into a latent space using a multi-layer perceptron, resulting in embeddings ${z_{g\in(p,s,v)}}_t^i\in\mathbb{R}^{D}$.

To better capture the fluctuating geometric cues, we design a hierarchical quantization strategy that discretizes the MLP-projected continuous embeddings. Specifically, for position, size, and velocity components, we apply a two-stage residual codebook to perform quantization.

\begin{equation}
E_{g \in (p, s, v)}^{(\ell)}=\{e_{g,1}^{(\ell)},\ldots,e_{g,K}^{(\ell)}\}\subset\mathbb{R}^{D}, \quad \ell=1,2,
\label{eq:codebook-level}
\end{equation}
where $D$ denotes the embedding dimension and $K$ represents the number of codewords per stage. We set $D=64$ and $K=64$ to balance representational capacity and computational efficiency. A detailed analysis of the hyperparameter design is provided in Sec.~\ref{sec:ablation}.

\begin{equation}
\left\{
\begin{aligned}
{k_1}_{g,t}^{i} &= \argmin_{k_1 \in \{1,\dots,K\}}
                 \big\lVert z_{g,t}^i - e_{g,{k_1}}^{(1)} \big\rVert_2,
&\quad q_{g,1} &= e_{g,{k_1}}^{(1)}, \\[2pt]
r_{g,t}^i &= z_{g,t}^i - q_{g,1}, \\[2pt]
{k_2}_{g,t}^{i} &= \argmin_{k_2 \in \{1,\dots,K\}}
                 \big\lVert r_{g,t}^i - e_{g,{k_2}}^{(2)} \big\rVert_2,
&\quad q_{g,2} &= e_{g,{k_2}}^{(2)}.
\end{aligned}
\right.
\label{eq:two-stage-quant}
\end{equation}

The residual formulation allows the second codebook to capture fine-grained geometric variations that persist after coarse quantization.
The final geometric embedding $ \tilde z_{g,t}^i$ is obtained by summing the coarse and fine codewords, producing a more detailed and robust representation~\cite{Lee_2022_CVPR}. 
The discrete embeddings $\tilde z_{p,t}^i$, $\tilde z_{s,t}^i$, and $\tilde z_{v,t}^i$ are then concatenated into a 3D-dimensional vector and linearly projected to $\tilde z_t^i \in \mathbb{R}^{d}$ to ensure compatibility with the visual features.

\subsection{Geometry-Aware Animal Tracking}
\label{sec:gat}
In addition to the object-level visual embedding \( f_t^i \) extracted in Sec.~\ref{sec:gvc}, we follow~\cite{Gao_2025_CVPR} and incorporate identity information by projecting a one-hot encoded ID index into an embedding \( l^m \) for trajectory \( m \). 
We further integrate the discretized geometric embedding \( \tilde{z}_t^i \) to jointly model appearance, identity, and geometric states. 
All embeddings \( f_t^i \), \( l^m \), and \( \tilde{z}_t^i \) are projected to dimension \( d \).
At frame \( t \), the historical representation of trajectory \( m \) over time window \( [t\!-\!T, t\!-\!1] \) is defined as:
\begin{equation}
F_{t-T:t-1}^m =
\left\{
\left(f_{t-T}^m, \tilde{z}_{t-T}^m, l^m\right), \dots,
\left(f_{t-1}^m, \tilde{z}_{t-1}^m, l^m\right)
\right\}.
\label{eq:frame_sequence}
\end{equation}

Each detected object in the current frame is represented as \( o_t^i = (f_t^i, \tilde{z}_t^i, l_{\text{init}}^i) \). A Transformer decoder performs cross-attention between the current object query \( o_t^i \) and historical trajectory features \( F_{t-T:t-1}^m \). The concatenation of visual, geometric, and identity embeddings preserves modality-specific information before attention interaction. By incorporating discretized geometric tokens, the attention mechanism becomes less sensitive to small coordinate perturbations, leading to more stable identity matching across frames. The attention output is combined with \( o_t^i \) and passed through a classification head for identity prediction.

\subsection{Training}
\label{sec:train}

\textbf{Loss Function.} Following prior works~\cite{Meinhardt_2022_CVPR,sun2020transtrack,zhang2021fairmot}, we formulate trajectory association as an identity classification task that includes an additional unknown category. Beyond the standard detection loss \( \mathcal{L}_{\text{det}} \) from the DETR, we incorporate an ID classification loss \( \mathcal{L}_{\text{id}} \) and a vector quantization loss \( \mathcal{L}_{\text{vq}} \) for training process. These components are integrated into a unified objective via a weighted sum, where each coefficient \( \lambda_i \) modulates the contribution of the corresponding loss term \( \mathcal{L}_i \). The overall training objective is formulated as:

\begin{equation}
\mathcal{L}_{\text{total}} =
\lambda_{\text{det}}\mathcal{L}_{\text{det}} +
\lambda_{\text{id}}\mathcal{L}_{\text{id}} +
\lambda_{\text{vq}}\mathcal{L}_{\text{vq}},
\label{eq:total_loss}
\end{equation}
where $\mathcal{L}_{\text{det}}$ consists of a classification loss, a Generalized IoU loss, and an L1 regression loss. For the ID classification loss $\mathcal{L}_{\text{id}}$, we adopt the standard cross-entropy formulation. 
The vector quantization loss $\mathcal{L}_{\text{vq}}$ is defined as $\| \text{sg}[z] - e \|_2^2 + \beta \| z - \text{sg}[e] \|_2^2$, where $\text{sg}[\cdot]$ denotes the stop-gradient operator. The first term updates the codewords $e$, while the second term encourages the encoder output $z$ to commit to its assigned codeword~\cite{yu2022vectorquantized,wang2024omnitokenizer}. 
To enable gradient backpropagation through the non-differentiable nearest-neighbor operation, we adopt the straight-through estimator (STE)~\cite{NIPS2017_7a98af17}.

\subsection{Inference}
\label{sec:inference}

During inference, object detections in the current frame lack explicit temporal correspondences with the previous frame, which prohibits direct velocity estimation via frame-to-frame differencing.
For each detection with normalized center $\mathbf{c}_t^i = (c_{x,t}^i, c_{y,t}^i)$, we estimate its velocity using a lightweight KNN-based regression over historical trajectory states. 
Specifically, we retrieve the $k$ nearest historical states $\mathcal{N}_k$ in spatial coordinate space and compute the estimated velocity as 
$\hat{\mathbf{v}}_t^i = \sum_{j \in \mathcal{N}_k} w_{ij} \mathbf{v}_j$, 
where the normalized weights are defined as
\begin{equation}
w_{ij} =
\frac{
\exp\!\left(-\|\mathbf{c}_t^i - \mathbf{c}_j\|_2^2 / \sigma_s^2\right)
\cdot
\exp\!\left(\cos(\Delta\phi_{ij}) / \sigma_d\right)
}{
\sum_{l \in \mathcal{N}_k}
\exp\!\left(-\|\mathbf{c}_t^i - \mathbf{c}_l\|_2^2 / \sigma_s^2\right)
\cdot
\exp\!\left(\cos(\Delta\phi_{il}) / \sigma_d\right)
}.
\end{equation}
Here $\mathbf{v}_j$ denotes historical velocity vectors and $\Delta\phi_{ij}$ denotes the angular difference between the motion direction of the current detection and the historical state $j$. 
The spatial kernel encourages proximity consistency, while the directional term favors motion-aligned trajectories. 

\vspace{-2mm}

\section{Experiments}

\vspace{-2mm}

\subsection{Datasets and Metrics}

\vspace{-2mm}

\textbf{Datasets.} To comprehensively evaluate multi-animal tracking, we select a set of challenging and representative benchmarks spanning diverse animal study scenarios. Animaltrack~\cite{zhang2023animaltrack} is a multi-class animal tracking dataset featuring complex inter-individual interactions, consisting of 58 sequences collected from wild animals across 10 terrestrial and aquatic categories. BFT~\cite{Zheng_2024_CVPR} is a dataset of 22 bird species with diverse morphological characteristics and irregular motion patterns, composed of 106 videos. BuckTales~\cite{NEURIPS2024_95286b5d} is a UAV-based animal tracking dataset featuring wild antelopes with highly similar visual appearances. Due to incomplete annotations in the full release, we evaluate on the subset with complete identity labels (12 videos, 680 trajectories). These benchmarks cover a broad spectrum of representative challenges in multi-animal tracking. This diverse setting enables a comprehensive evaluation of HieDG in terms of both performance and robustness across complex animal tracking scenarios.

To further assess the generalization of our study beyond animal-specific scenarios, we conduct experiments on two widely adopted multi-person tracking benchmarks, DanceTrack~\cite{Sun_2022_CVPR} and SportsMOT~\cite{Cui_2023_ICCV}. Both datasets exhibit extreme intra-class similarity and complex motion patterns, posing substantial challenges for identity association. 

\noindent\textbf{Metrics.} We adopt Higher Order Tracking Accuracy (HOTA)~\cite{luiten2021hota} to jointly evaluate detection (DetA) and association (AssA) performance. MOTA~\cite{bernardin2008evaluating} and IDF1~\cite{ristani2016performance} are also reported to cover complementary tracking aspects. Since our method primarily targets identity association, we place particular attention on HOTA, AssA, and IDF1, while reporting MOTA for completeness.

\vspace{-3mm}

\subsection{Implementation Details} 

\vspace{-2mm}
Given its simplicity and generality, we adopt MOTIP~\cite{Gao_2025_CVPR} as the baseline for this study. Specifically, we employ a ResNet-based~\cite{he2016deep} Deformable DETR~\cite{zhu2021deformable} as the detector in the proposed HieDG framework. To maintain consistency with prior work, we initialize the model using COCO~\cite{lin2014microsoft} pretrained weights. Furthermore, we set the dimension of each codebook and each geometric component to $D=64$, while the hidden dimension of the quantized geometry embedding $\tilde{\mathbf{z}}$ is aligned with the Deformable DETR backbone, i.e., $C=256$.

\begin{table*}[t]
  \centering
  \setlength{\tabcolsep}{3.2pt}
  \renewcommand{\arraystretch}{1.08}

  \caption{
  Performance comparison on the AnimalTrack~\cite{zhang2023animaltrack} and BFT~\cite{Zheng_2024_CVPR} test sets.
  * denotes the use of continuous geometric cues.
  The best and second-best performance are highlighted in \textbf{bold} and \underline{underline}, respectively.
  }
  \label{tab:animaltrack_bft_results}

  \vspace{2pt}

  \resizebox{\linewidth}{!}{
  \begin{tabular}{@{}>{\raggedright\arraybackslash}p{3.0cm}|ccccc|ccccc@{}}
    \toprule[1.2pt]
    \multirow{2}{*}{Methods} 
    & \multicolumn{5}{c|}{AnimalTrack~\cite{zhang2023animaltrack}} 
    & \multicolumn{5}{c}{BFT~\cite{Zheng_2024_CVPR}} \\
    \cmidrule(lr){2-6}\cmidrule(lr){7-11}
    & HOTA & DetA & AssA & MOTA & IDF1
    & HOTA & DetA & AssA & MOTA & IDF1 \\
    \midrule[0.6pt]

    \multicolumn{11}{@{}l@{}}{\textbf{\textit{Heuristic-based}}} \\
    JDE~\cite{JDE}                  & 28.7 & 40.6 & 20.3 & 28.3 & 33.2 & 30.7 & 40.9 & 23.4 & 35.4 & 37.4 \\
    CSTrack~\cite{CSTrack}          & --   & --   & --   & --   & --   & 33.2 & 47.0 & 23.7 & 46.7 & 34.5 \\
    FairMOT~\cite{zhang2021fairmot} & 30.6 & --   & --   & 29.0 & 38.8 & 40.2 & 53.3 & 28.2 & 56.0 & 41.8 \\
    TADAM~\cite{TADAM}              & 32.5 & --   & --   & 36.5 & 37.2 & --   & --   & --   & --   & --   \\
    DeepSORT~\cite{10222576}        & 36.4 & 41.2 & 32.1 & 41.0 & 35.2 & --   & --   & --   & --   & --   \\
    SORT~\cite{7533003}             & 42.5 & 48.8 & 37.0 & 54.8 & 49.1 & 61.2 & 60.6 & 62.3 & 75.5 & 77.2 \\
    IOUTrack~\cite{8078516}         & 41.6 & --   & --   & 55.7 & 45.7 & --   & --   & --   & --   & --   \\
    Tracktor++~\cite{Bergmann_2019_ICCV} & 44.2 & -- & -- & 55.2 & 51.0 & -- & -- & -- & -- & -- \\
    QDtrack~\cite{Pang_2021_CVPR}   & 47.0 & --   & --   & 55.7 & 56.3 & --   & --   & --   & --   & --   \\
    TransCenter~\cite{transcenter}  & --   & --   & --   & --   & --   & 60.0 & 66.0 & 61.1 & 74.1 & 72.4 \\
    CenterTrack~\cite{centertrack}  & --   & --   & --   & --   & --   & 56.2 & 58.5 & 54.0 & 60.2 & 61.0 \\
    ByteTrack~\cite{ByteTrack}      & 48.1 & 46.9 & 49.3 & 56.6 & 55.5 & 62.5 & 61.2 & 64.1 & 77.2 & 82.3 \\
    OC-SORT~\cite{Cao_2023_CVPR}    & --   & --   & --   & --   & --   & 66.8 & 65.4 & 68.7 & 77.1 & 79.3 \\

    \midrule[0.6pt]
    \multicolumn{11}{@{}l@{}}{\textbf{\textit{Query-based}}} \\
    TrackFormer~\cite{Meinhardt_2022_CVPR} & 31.0 & -- & -- & 20.4 & 36.5 & 63.3 & 66.0 & 61.1 & 74.1 & 72.4 \\
    TransTrack~\cite{sun2020transtrack}    & 45.4 & -- & -- & 48.3 & 53.4 & 65.1 & 68.5 & 61.8 & 74.9 & 73.1 \\
    MOTR~\cite{zeng2022motr}               & 49.5 & 51.3 & 47.7 & 57.2 & 54.1 & 64.2 & 64.4 & 64.7 & 74.1 & 75.2 \\
    MeMOTR~\cite{Gao_2023_ICCV}            & 52.2 & 54.3 & 50.1 & 62.1 & 58.6 & --   & --   & --   & --   & --   \\
    MOTIP~\cite{Gao_2025_CVPR}             & 54.1 & 54.2 & 55.0 & 61.9 & 61.7 & 69.2 & \textbf{71.1} & 67.3 & \textbf{78.6} & \underline{80.1} \\
    CO-MOT~\cite{yan2025comot}             & \underline{55.3} & \textbf{55.1} & \underline{55.5} & \textbf{62.8} & \underline{62.1}
                                           & 69.0 & 70.5 & 67.7 & \underline{78.3} & 79.0 \\

    \midrule[0.6pt]
    \multicolumn{11}{@{}l@{}}{\textbf{\textit{Ours}}} \\
    HieDG* & 54.6 & \underline{55.0} & 54.2 & 62.4 & 61.8 & \underline{69.5} & 70.4 & \underline{68.8} & 77.2 & 79.8 \\
    HieDG  & \textbf{56.2} & 54.9 & \textbf{58.4} & \underline{62.5} & \textbf{64.4} 
           & \textbf{71.3} & \underline{70.6} & \textbf{72.2} & 77.4 & \textbf{82.5} \\
    \bottomrule[1.2pt]
  \end{tabular}
  }

\end{table*}

Following prior work~\cite{Gao_2025_CVPR,motrv2,Gao_2023_ICCV}, we apply standard image-level augmentations, including random resizing, horizontal flipping, and color jittering, for an apples-to-apples comparison. To improve robustness to estimation noise at inference, we apply a 0.3 dropout to velocity embeddings during training, which regularizes the model against potential motion uncertainty. All experiments are implemented in PyTorch and conducted on NVIDIA Tesla V100 GPUs. We use 4 GPUs for training on the AnimalTrack, DanceTrack, and SportsMOT datasets, and 2 GPUs for BFT and BuckTales.

We adopt the AdamW optimizer with a weight decay of \(5 \times 10^{-4}\) and an initial learning rate of \(10^{-4}\). A linear warmup is applied for the first epoch to stabilize learning rate adjustment. The loss weight coefficients are set to \(\lambda_{\text{det}} = 1.0\), \(\lambda_{\text{id}} = 2.0\), and \(\lambda_{\text{vq}} = 0.1\).

\vspace{-5mm}

\subsection{Comparisons with State-of-the-art Methods}
\label{sec:comparisons}

\vspace{-3mm}

We evaluate the proposed HieDG in comparison with representative heuristic-based and query-based trackers on the AnimalTrack~\cite{zhang2023animaltrack}, BFT~\cite{Zheng_2024_CVPR}, BuckTales~\cite{NEURIPS2024_95286b5d}, DanceTrack~\cite{Sun_2022_CVPR}, and SportsMOT~\cite{Cui_2023_ICCV} benchmarks. The quantitative results are illustrated in Tab.~\ref{tab:animaltrack_bft_results}, Tab.~\ref{tab:buck_results}, and Tab.~\ref{tab:dancetrack_sportsmot_results}. Since association mechanisms are inherently coupled with detection backbones, we report results following each method’s official evaluation protocols and configurations.

\noindent\textbf{AnimalTrack and BFT.} We evaluate HieDG on standard multi-animal tracking datasets in AnimalTrack and BFT across terrestrial, underwater, and aerial environments, where individuals frequently exhibit uniform appearance, high density, and irregular movements. Heuristic-based methods (e.g., SORT~\cite{7533003}, DeepSORT~\cite{10222576}, and OC-SORT~\cite{Cao_2023_CVPR}) rely on handcrafted motion and association rules. Although OC-SORT performs better at handling abrupt nonlinear motion in BFT for bird tracking, such strategies remain limited under complex multi-animal dynamics. Query-based approaches demonstrate greater adaptability, yet HieDG consistently outperforms existing methods and achieves state-of-the-art association performance, yielding clear gains in HOTA, AssA, and IDF1. A detailed comparison reveals a consistent trend: directly incorporating continuous geometric representation (HieDG$^{*}$) degrades association accuracy relative to the baseline, whereas discretizing geometric embedding consistently yields performance gains. This contrast suggests that continuous geometry can be unreliable in crowded and appearance-ambiguous scenarios, where it may introduce substantial noise. By contrast, discretization converts raw geometry into stable structural tokens, thereby enhancing association robustness. These results suggest that discretized geometric modeling provides improved robustness in complex multi-animal tracking scenarios.

\vspace{-5mm}

\noindent
\begin{wraptable}{l}{0.50\linewidth}
  \vspace{-10pt}
  \centering

  \caption{
  Performance comparison on the BuckTales~\cite{NEURIPS2024_95286b5d} test set.
  * denotes the use of continuous geometric cues.
  Best and second-best results are highlighted in \textbf{bold} and \underline{underline}, respectively.
  }
  \label{tab:buck_results}

  \vspace{2pt}

  \setlength{\tabcolsep}{3.2pt}
  \renewcommand{\arraystretch}{1.08}

  \resizebox{\linewidth}{!}{
  \begin{tabular}{@{}>{\raggedright\arraybackslash}p{2.6cm}|ccccc@{}}
    \toprule[1.2pt]
    Methods & HOTA & DetA & AssA & MOTA & IDF1 \\
    \midrule[0.6pt]

    \multicolumn{6}{@{}l@{}}{\textbf{\textit{Heuristic-based}}} \\

    SORT~\cite{7533003}
      & 30.1 & 40.7 & 24.1 & 42.9 & 47.5 \\

    DeepSORT~\cite{10222576}
      & 9.62 & 25.6 & 3.8 & 16.3 & 56.5 \\

    ByteTrack~\cite{ByteTrack}
      & \underline{49.8} & \textbf{47.1} & \underline{52.6}
      & \textbf{49.6} & \underline{64.0} \\

    OC-SORT~\cite{Cao_2023_CVPR}
      & 42.4 & 45.1 & 39.8 & \underline{47.3} & 57.9 \\

    \midrule[0.6pt]

    \multicolumn{6}{@{}l@{}}{\textbf{\textit{Query-based}}} \\

    TrackFormer~\cite{Meinhardt_2022_CVPR}
      & 41.7 & \underline{45.8} & 37.9 & 41.4 & 57.2 \\

    MOTR~\cite{zeng2022motr}
      & 42.5 & 44.0 & 41.1 & 41.8 & 58.6 \\

    MOTIP~\cite{Gao_2025_CVPR}
      & 45.6 & 42.2 & 50.4 & 43.9 & 61.8 \\

    CO-MOT~\cite{yan2025comot}
      & 47.2 & 43.6 & 51.3 & 45.9 & 62.3 \\

    \midrule[0.6pt]

    \multicolumn{6}{@{}l@{}}{\textbf{\textit{Ours}}} \\

    HieDG*
      & 41.3 & 43.2 & 39.4 & 44.9 & 56.6 \\

    HieDG
      & \textbf{52.4} & 42.9 & \textbf{64.0}
      & 44.6 & \textbf{69.7} \\

    \bottomrule[1.2pt]
  \end{tabular}
  }

  \vspace{-12pt}

\end{wraptable}

\noindent\textbf{BuckTales.}
UAV-based remote sensing, widely used in wildlife monitoring, introduces additional challenges for multi-animal tracking. Aerial viewpoints yield small, blurry targets, while frequent camera shake and abrupt viewpoint shifts destabilize both appearance and spatial relationships, making identity association significantly more difficult. As shown in Tab.~\ref{tab:buck_results}, ByteTrack achieves competitive performance in this UAV setting, suggesting that motion-driven heuristics can remain effective when appearance cues collapse. Nevertheless, HieDG achieves the best overall association performance. The improvements in HOTA, AssA, and IDF1 indicate stronger identity consistency under severe geometric perturbations. These findings are consistent with the observations on AnimalTrack and BFT, further demonstrating the robustness of hierarchical geometric modeling in extreme aerial tracking scenarios.

\begin{table*}[t]
  \centering
  \setlength{\tabcolsep}{3.2pt}
  \renewcommand{\arraystretch}{1.08}

  \caption{
  Performance comparison on the DanceTrack~\cite{Sun_2022_CVPR} and SportsMOT~\cite{Cui_2023_ICCV} test sets.
  * denotes the use of continuous geometric cues.
  The best and second-best performance are highlighted in \textbf{bold} and \underline{underline}, respectively.
  }
  \label{tab:dancetrack_sportsmot_results}

  \vspace{2pt}

  \resizebox{\linewidth}{!}{
  \begin{tabular}{@{}>{\raggedright\arraybackslash}p{3.0cm}|ccccc|ccccc@{}}
    \toprule[1.2pt]
    \multirow{2}{*}{Methods} 
    & \multicolumn{5}{c|}{DanceTrack~\cite{Sun_2022_CVPR}} 
    & \multicolumn{5}{c}{SportsMOT~\cite{Cui_2023_ICCV}} \\
    \cmidrule(lr){2-6}\cmidrule(lr){7-11}
    & HOTA & DetA & AssA & MOTA & IDF1
    & HOTA & DetA & AssA & MOTA & IDF1 \\
    \midrule[0.6pt]

    \multicolumn{11}{@{}l@{}}{\textbf{\textit{Heuristic-based}}} \\
    FairMOT~\cite{zhang2021fairmot}  & 39.7 & 66.7 & 23.8 & 82.2 & 40.8 & 49.3 & 70.2 & 34.7 & 86.4 & 53.5 \\
    QDtrack~\cite{Pang_2021_CVPR}  & 54.2 & 80.1 & 36.8 & 87.7 & 50.4 & 60.4 & 77.5 & 47.2 & 90.1 & 62.3 \\
    ByteTrack~\cite{ByteTrack} & 47.7 & 71.0 & 32.1 & 89.6 & 53.9 & 62.1 & 76.5 & 50.5 & \textbf{93.4} & 69.1 \\
    OC-SORT~\cite{Cao_2023_CVPR} & 55.1 & 80.3 & 38.3 & \textbf{92.0} & 54.6 & 68.1 & \textbf{84.8} & 54.8 & \textbf{93.4} & 68.0 \\

    \midrule[0.6pt]
    \multicolumn{11}{@{}l@{}}{\textbf{\textit{Query-based}}} \\

    TrackFormer~\cite{Meinhardt_2022_CVPR} & -- & -- & -- & -- & -- & 63.3 & 66.0 & 61.1 & 74.1 & 72.4 \\

    MeMOTR~\cite{Gao_2023_ICCV} & 63.4 & 77.0 & 52.3 & 85.4 & 65.5 & 68.8 & 82.0 & 57.8 & 90.2 & 69.9 \\

    TransTrack~\cite{sun2020transtrack} & 45.5 & 75.9 & 27.5 & 88.4 & 45.2 & 68.9 & 82.7 & 57.5 & \underline{92.6} & 71.5 \\

    CO-MOT\cite{yan2025comot}  & 65.3 & 80.1 & 53.5 & 89.3 & 66.5 & -- & -- & -- & -- & -- \\

    MOTIP~\cite{Gao_2025_CVPR}  & \underline{69.6} & 80.4 & \underline{60.4} & 90.6 & \underline{74.7}
          & \underline{72.6} & \underline{83.5} & \underline{63.2} & 92.4 & \underline{77.1} \\

    \midrule[0.6pt]
    \multicolumn{11}{@{}l@{}}{\textbf{\textit{Ours}}} \\

    HieDG* & 68.5 & \textbf{80.8} & 60.0 & \underline{91.4} & 73.1
           & 71.4 & 82.7 & 59.6 & 92.5 & 71.7 \\

    HieDG  & \textbf{70.3} & \underline{80.7} & \textbf{61.8} & 90.9 & \textbf{75.4}
           & \textbf{72.8} & 83.4 & \textbf{63.9} & 92.1 & \textbf{77.5} \\

    \bottomrule[1.2pt]
  \end{tabular}
  }
\end{table*}

\noindent\textbf{DanceTrack and SportsMOT.}
To evaluate generalization in generic low-discriminability multi-person tracking scenarios, we report results on DanceTrack and SportsMOT in Tab.~\ref{tab:dancetrack_sportsmot_results}. HieDG achieves superior association performance across both benchmarks. The consistent gains in AssA and IDF1 demonstrate improved identity preservation in crowded and appearance-ambiguous scenes. These results indicate that the proposed geometry-guided modeling strategy generalizes beyond animal-specific tracking and remains effective in broader multi-object tracking settings.

\subsection{Ablations}
\label{sec:ablation}
We conduct ablation studies on the BFT~\cite{Zheng_2024_CVPR} dataset, which exhibits dense interactions, strong appearance similarity, and irregular motion patterns. 
These characteristics make BFT a suitable testbed for validating the effectiveness of each component in our framework under low-discriminability conditions. For clarity, we denote HieDG* as the variant using continuous geometric embeddings without quantization.

\noindent\textbf{Geometric Component.} Heuristic-based methods typically leverage geometric cues, such as position and scale, from detectors to guide tracking. To capture the fast yet coherent motion patterns in animal scenarios, we model velocity based on inter-frame position differences. As shown in Tab.~\ref{tab:ablation_geom_components}, each geometric cue independently improves association performance, while combining all three yields the highest performance gain. These results demonstrate that complementary geometric cues provide additive benefits for identity association, and that jointly modeling position, scale, and velocity yields the most stable cross-frame matching.

\begin{table}[t]
\centering

\begin{minipage}[t]{0.485\linewidth}
\centering
\vspace{0pt}

\caption{Impact of different geometric components. The best performance is shown in \textbf{bold}, and the gray background is the choice for our final experiment.}
\label{tab:ablation_geom_components}
\vspace{2pt}

\setlength{\tabcolsep}{3.5pt}
\renewcommand{\arraystretch}{1.12}
\resizebox{\linewidth}{!}{
\begin{tabular}{@{}ccc|ccccc@{}}
  \toprule[1.2pt]
  Pos. & Scale & Vel. & HOTA & DetA & AssA & MOTA & IDF1 \\
  \midrule[0.7pt]
        &       &       & 69.2 & 71.1 & 67.3 & 78.6 & 80.1 \\
  \cmark&       &       & 70.6 & 70.5 & 71.0 & 76.9 & 81.6 \\
        & \cmark&       & 70.1 & 71.3 & 68.9 & 78.9 & 81.3 \\
        &       & \cmark& 69.7 & \textbf{71.8} & 68.0 & \textbf{79.1} & 80.9 \\
  \rowcolor{gray!15}
  \cmark& \cmark& \cmark& \textbf{71.3} & 70.6 & \textbf{72.2} & 77.4 & \textbf{82.5} \\
  \bottomrule[1.2pt]
\end{tabular}
}
\end{minipage}
\hfill
\begin{minipage}[t]{0.485\linewidth}
\centering
\vspace{0pt}

\caption{Impact of hierarchical geometric quantization strategies. The best performance is shown in \textbf{bold}, and the gray background is the choice for our final experiment.}
\label{tab:ablation_encoding}
\vspace{2pt}

\setlength{\tabcolsep}{3.5pt}
\renewcommand{\arraystretch}{1.10}
\resizebox{\linewidth}{!}{
\begin{tabular}{@{}l|ccc|cc@{}}
  \toprule[1.2pt]
  Geo. Emb. & HOTA & AssA & IDF1 & Para. (K) & FLOPs (K)\\
  \midrule[0.6pt]
  Cont. Enc. & 69.5 & 68.8 & 79.8 & +\textbf{199.2} & +198.1 \\
  FixVQ      & 69.8 & 69.1 & 80.2 & +49.9 & +99.0 \\
  1-Stage VQ & 70.5 & 68.9 & 81.3 & +99.1  & +197.4 \\
  \rowcolor{gray!15}
  2-Stage VQ & 71.3 & \textbf{72.2} & 82.5 & +148.3 & +295.6 \\
  3-Stage VQ & \textbf{71.5} & \textbf{72.2} & \textbf{82.7} & +197.4 & +\textbf{394.0} \\
  \bottomrule[1.2pt]
\end{tabular}
}
\end{minipage}

\end{table}

\begin{figure*}[h]
    \centering
    \begin{subfigure}[b]{1\textwidth}
        \centering
        \includegraphics[width=\linewidth]{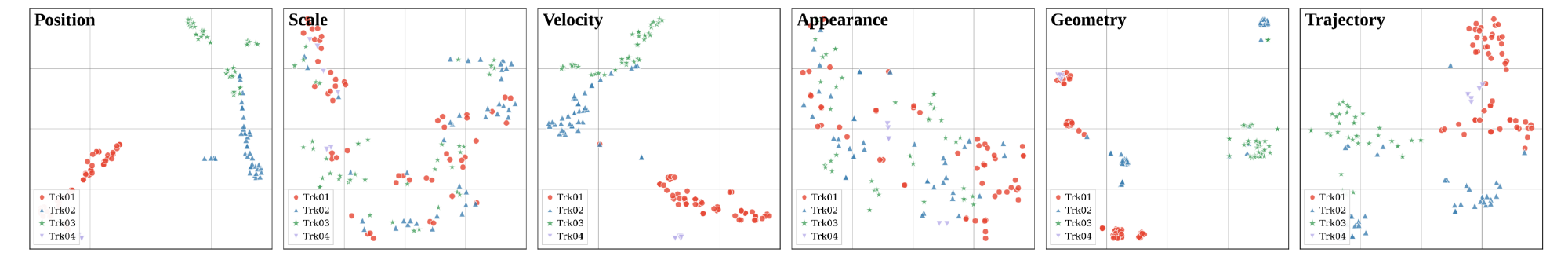}
        \caption{t-SNE~\cite{maaten2008visualizing} visualization of all frames in Fa5001 (BFT).}
    \end{subfigure}

    \begin{subfigure}[b]{1\textwidth}
        \centering
        \includegraphics[width=\linewidth]{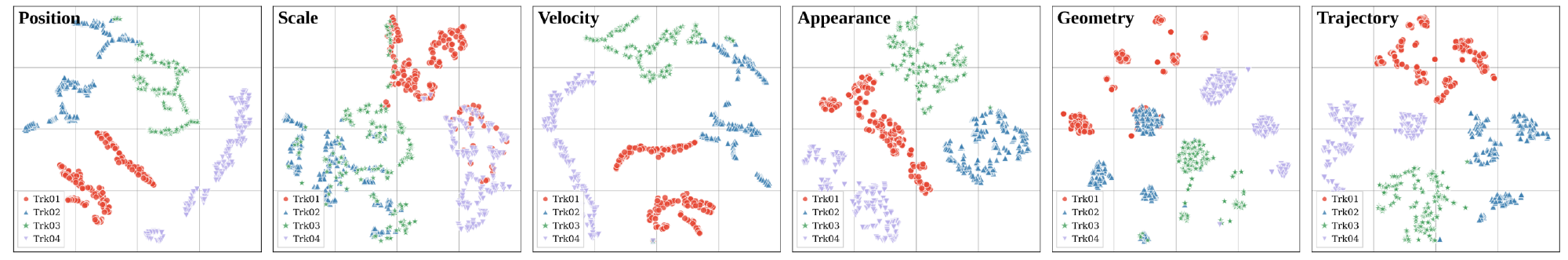}
        \caption{t-SNE~\cite{maaten2008visualizing} visualization of all frames in An1005 (BFT).}
    \end{subfigure}

    \caption{t-SNE~\cite{maaten2008visualizing} visualization of track embeddings across all frames in the BFT dataset. Different IDs are marked by distinct colors and shapes. }
    \label{fig:tsne}
\end{figure*}

We visualize the t-SNE~\cite{maaten2008visualizing} of trajectory embeddings for all frames in sequence Fa5001 and An1005, as shown in Fig.~\ref{fig:tsne}, where each color and shape denotes a unique ID. The first three columns illustrate embeddings derived from individual geometric components (position, scale, and velocity), whereas the last three columns show appearance features, discretized geometry embeddings, and the fused trajectory representation. While appearance features show moderate separability, the quantized geometric embeddings exhibit tighter intra-ID clustering compared to continuous appearance features. 
The fused embeddings further improve inter-ID separation, suggesting that discretized geometry stabilizes identity representation across frames.

\noindent\textbf{Geometric Vector Quantization.} Discretizing geometric cues into stable representations significantly enhances tracking performance. To evaluate our quantization design, we compare it with fixed-bin quantization and explore different depths of quantization. As shown in Tab.~\ref{tab:ablation_encoding}, fixed-bin quantization yields only limited gains, while hierarchical residual quantization consistently improves association performance. Although extending from two to three stages slightly increases HOTA and IDF1, the improvement is marginal relative to the additional 49.1K parameters and 98.4K FLOPs. This disproportionate increase in computational cost does not justify the minor accuracy gain. Therefore, we adopt the two-stage design as a favorable trade-off between performance and efficiency.

\begin{table}[t]
\centering

\begin{minipage}[t]{0.485\linewidth}
\centering
\vspace{0pt}

\caption{Impact of different codebook embedding dimensions. The best performance is shown in \textbf{bold}, and the gray background is the choice for our final experiment.}
\label{tab:ablation_codebook_dim}
\vspace{2pt}

\setlength{\tabcolsep}{3pt}
\renewcommand{\arraystretch}{1.12}
\resizebox{\linewidth}{!}{
\begin{tabular}{@{}c|ccc|cc@{}}
  \toprule[1.2pt]
  Code. Dim. & HOTA & AssA & IDF1 & Para. (K) & FLOPs (K) \\
  \midrule[0.6pt]
  16  & 70.2 & 70.3 & 80.7 & +74.6 & +148.2 \\
  32  & 70.8 & 71.7 & 81.6 & +99.1 & +197.4 \\
  \rowcolor{gray!15}
  64  & 71.3 & \textbf{72.2} & 82.5 & +148.3 & +295.6 \\
  128 & \textbf{71.4} & \textbf{72.2} & \textbf{82.7} & +\textbf{246.6} & +\textbf{492.3} \\
  \bottomrule[1.2pt]
\end{tabular}
}
\end{minipage}
\hfill
\begin{minipage}[t]{0.485\linewidth}
\centering
\vspace{0pt}

\caption{Impact of Hungarian matching and polar coordinate representation. The best performance is shown in \textbf{bold}, and the gray background is the choice for our final experiment.}
\label{tab:ablation_hungarian_polar}
\vspace{2pt}

\setlength{\tabcolsep}{3.5pt}
\renewcommand{\arraystretch}{1.12}
\resizebox{\linewidth}{!}{
\begin{tabular}{@{}c|c|ccccc@{}}
  \toprule[1.2pt]
  Hung. & Polar vel. & HOTA & DetA & AssA & MOTA & IDF1 \\
  \midrule[0.7pt]
        &            & 69.8 & \textbf{71.3} & 68.3 & \textbf{78.8} & 80.7 \\
  \cmark&            & 69.9 & 71.1 & 68.7 & 78.5 & 81.0 \\
  \rowcolor{gray!15}
        & \cmark     & \textbf{71.3} & 70.6 & \textbf{72.2} & 77.4 & \textbf{82.5} \\
  \cmark& \cmark     & 71.1 & 70.7 & 71.5 & 77.9 & 81.9 \\
  \bottomrule[1.2pt]
\end{tabular}
}
\end{minipage}

\end{table}

To assess the impact of codebook size, we set equal dimensions for the position, scale, and velocity codebooks to ensure a fair comparison. As shown in Tab.~\ref{tab:ablation_codebook_dim}, increasing the codebook size to 64 substantially improves association performance with moderate computational overhead. However, further scaling to 128 offers marginal gains while introducing significant additional cost. These results suggest that our two-stage residual quantization sufficiently captures fluctuating geometry in animal scenarios, rendering larger codebooks less beneficial.

\noindent\textbf{Hungarian Algorithm and Velocity Representation.} Inspired by~\cite{Gao_2025_CVPR}, although the Hungarian algorithm is widely used for global ID assignment, our association strategy (detailed in Sec.~\ref{sec:gat}) is formulated independently of this matching mechanism. To enhance robustness to low-speed jitter and rapid directional changes common in animal scenarios, we model velocity in polar coordinates. The ablation results in Tab.~\ref{tab:ablation_hungarian_polar} show that incorporating Hungarian matching brings negligible improvement, indicating that the ID classification-based formulation already provides effective global assignment within the learned framework. In contrast, modeling velocity in polar coordinates consistently improves AssA and IDF1, suggesting that decoupling magnitude and direction provides a more stable motion representation under irregular animal dynamics.

\section{Conclusion}

We propose HieDG, a hierarchical discrete geometry-guided framework for multi-animal tracking. 
By reformulating continuous geometric cues into structured discrete embeddings through residual quantization, HieDG stabilizes cross-frame identity association within attention-based tracking architectures. 
Extensive experiments demonstrate consistent improvements across diverse multi-animal benchmarks, with competitive generalization to generic tracking datasets. 
Overall, these findings suggest that discretized geometric representations can improve the stability of attention-based identity modeling under low-discriminability and noisy motion conditions.

\section{Acknowledgement}

This work was supported by the National Key Research and Development Program of China under Grant 2023YFC3208303; the Brain Science and Brain-like Intelligence Technology-National Science and Technology Major Project under Grants 2022ZD0211900 and 2022ZD0211902; the National Natural Science Foundation of China (NSFC) under Grant 82471245 from Hang Zhou, Grant 12301651 from Weifu Li, and Grant 62273347 from Yifan Zhang; the Shenzhen Natural Science Foundation under Grant JCYJ20241202130204005; the Guangdong Basic and Applied Basic Research Foundation under Grant 2025A1515010844; and the Characteristic Innovation Project of Guangdong Higher Education Institutions (Natural Science) under Grant 2024KTSCX027.

\clearpage

\clearpage
\setcounter{page}{1}

\begin{center}
{\Large \bfseries Supplementary Material}\\[2mm]
{\large HieDG: A Hierarchical Discrete Geometry-Guided Framework for Multi-Animal Tracking}
\end{center}

\appendix

\section{Overview}
In this supplementary material, we primarily:
\begin{enumerate}
    \item State additional experimental details, in Sec.~\ref{experimental_details_supple}.
    \item Provide the pseudo-algorithm of the hierarchical geometry quantization process, in Sec.~\ref{PseudoAlgorithm}.
    \item Present additional visualization results across different benchmarks, in Appendix~\ref{sec:visualization}.
    \item Discuss the strengths, limitations, and potential future directions of this work in Sec.~\ref{sec:Discussion}. 
\end{enumerate}

\section{Experimental Details}
\label{experimental_details_supple}

\subsection{Dataset Details}
\label{sec:dataset_details}

Due to space constraints in the main paper, only a brief summary of the datasets was provided. Here we supplement more detailed information about the three benchmarks used in our experiments. AnimalTrack~\cite{zhang2023animaltrack} contains 402.0K annotated instances across 10 animal categories, including aquatic and terrestrial species such as dolphins, deer, and geese. BFT~\cite{Zheng_2024_CVPR} focuses on aerial bird tracking and includes 85.8K annotations covering 22 bird species across the training, validation, and test splits. In contrast, BuckTales~\cite{NEURIPS2024_95286b5d} is a UAV-based dataset centered on buck tracking. Although it contains fewer image frames, it includes 853.7K annotated instances, indicating a much higher object density and introducing greater tracking challenges due to increased occlusions and complex interactions. The overall statistics are summarized in Tab.~\ref{tab:dataset_stats}.

\begin{table}[h]
\centering
\small
\setlength{\tabcolsep}{3pt}
\renewcommand{\arraystretch}{1.12}
\begin{tabular}{@{}l|c|ccc|c@{}}
\toprule[1.2pt]
Dataset & \#Cls & Train imgs & Val imgs & Test imgs & \#Obj \\
\midrule[0.6pt]
AnimalTrack~\cite{zhang2023animaltrack}
& 10 
& 15.3K 
& 3.8K 
& 4.0K 
& 402.0K \\
BFT~\cite{Zheng_2024_CVPR}
& 22 
& 8.4K 
& 5.0K 
& 5.8K 
& 85.8K \\
BuckTales~\cite{NEURIPS2024_95286b5d}
& 1 
& 9.0K 
& 3.5K 
& 3.5K 
& 853.7K \\
\bottomrule[1.2pt]
\end{tabular}
\caption{
Statistics of the three multi-animal tracking benchmarks used in our study. The table summarizes the number of categories, the split of training/validation/test images, and the total number of annotated instances.
}
\label{tab:dataset_stats}
\end{table}

\begin{algorithm}[t]
\caption{Hierarchical Geometry Quantization in HieDG}
\label{alg:hgq_full}
\begin{algorithmic}[1]

\REQUIRE Normalized geometry $g_t^i = (g_{p,t}^i, g_{s,t}^i, g_{v,t}^i)$
\REQUIRE MLPs $\phi_g(\cdot)$ for $g \in \{p,s,v\}$; two-stage codebooks $E_g^{(1)},E_g^{(2)}$; fusion projection $(W,b)$; VQ weight $\beta$
\ENSURE Discrete embeddings $\tilde z_t^i$ and VQ loss $\mathcal{L}_{\mathrm{vq}}$

\STATE $\mathcal{L}_{\mathrm{vq}} \gets 0$

\FOR{$t = 1,\dots,T$}                      
  \FOR{each object $i$ in frame $t$}       

    \STATE $g_{p,t}^i \gets (c_{x,t}^i/W,\; c_{y,t}^i/H)$
           \hfill \textit{normalized position}

    \STATE $g_{s,t}^i \gets (w_t^i/W,\; h_t^i/H)$
           \hfill \textit{normalized scale}

    \STATE $g_{v,t}^i \gets (\rho_t^i,\; \theta_t^i)$
           \hfill \textit{polar velocity}

    \FOR{$g \in \{p,s,v\}$}
      \STATE \textit{Process each geometric component.}

      \STATE $z_{g,t}^i \gets \phi_g(g_{g,t}^i)$
             \hfill \textit{MLP projection}

      \STATE $k_{1,g,t}^i \gets
      \arg\min_{k \in \{1,\dots,K\}}
      \bigl\| z_{g,t}^i - e_{g,k}^{(1)} \bigr\|_2$
             \hfill \textit{stage-1 index}

      \STATE $q_{g,1,t}^i \gets e_{g,k_{1,g,t}^i}^{(1)}$
             \hfill \textit{coarse codeword}

      \STATE $r_{g,t}^i \gets z_{g,t}^i - q_{g,1,t}^i$
             \hfill \textit{residual}

      \STATE $k_{2,g,t}^i \gets
      \arg\min_{k \in \{1,\dots,K\}}
      \bigl\| r_{g,t}^i - e_{g,k}^{(2)} \bigr\|_2$
             \hfill \textit{stage-2 index}

      \STATE $q_{g,2,t}^i \gets e_{g,k_{2,g,t}^i}^{(2)}$
             \hfill \textit{residual codeword}

      \STATE $\tilde z_{g,t}^i \gets q_{g,1,t}^i + q_{g,2,t}^i$
             \hfill \textit{final quantized component}

      \STATE $\mathcal{L}_{\mathrm{vq}} \gets \mathcal{L}_{\mathrm{vq}}
      + \bigl\|\mathrm{sg}[z_{g,t}^i] - \tilde z_{g,t}^i\bigr\|_2^2
      + \beta\,\bigl\|z_{g,t}^i - \mathrm{sg}[\tilde z_{g,t}^i]\bigr\|_2^2$
             \hfill \textit{update VQ loss}

    \ENDFOR

    \STATE $\tilde z_t^i \gets
    W\bigl[\,\tilde z_{p,t}^i;\,\tilde z_{s,t}^i;\,\tilde z_{v,t}^i\,\bigr] + b$
           \hfill \textit{fuse geometry}

  \ENDFOR
\ENDFOR

\STATE \textbf{return} $\{\tilde z_t^i\},\; \mathcal{L}_{\mathrm{vq}}$

\end{algorithmic}
\end{algorithm}

\subsection{Training}
\label{sec:training}

Building on these dataset characteristics, we next describe key aspects of our training strategy. Since the model is initialized with COCO-pretrained DETR weights, the visual features are more reliable than the geometric and ID embeddings at the start of training. Without additional regularization, the model tends to underuse or even discard the parameters linked to geometry and ID cues. To address this, we adopt a U-shaped curriculum. During the first 5\% of training, we use a standard setup so the model can establish basic tracking ability with strong visual guidance. Between 5\% and 35\%, we apply a 0.35 dropout rate to the visual embeddings in the HieDG framework, which shifts the learning focus toward geometric and ID information. To maintain consistency between training and inference, we reduce the visual embedding dropout to 0.1 to recover appearance details. This schedule follows a simple-to-complex trajectory: learn the basic pattern first, strengthen geometric reasoning next, and finally reintegrate appearance cues, resulting in more stable optimization and stronger generalization.

\section{Pseudo Algorithm}
\label{PseudoAlgorithm}

We present the pseudo-algorithm of our hierarchical geometry quantization in \cref{alg:hgq_full}. In this process, we first normalize the positions and scales of the detected $bbox$ to handle the varying resolutions and aspect ratios across videos. Given the complex motion patterns common in animal scenes, we further convert the frame-to-frame Cartesian displacements into a polar-coordinate velocity representation. We then employ three separate MLPs to map the normalized position, size, and velocity cues into their respective embeddings.

Moreover, we introduce a two-stage codebook for geometric quantization. We first use a nearest-neighbor search to match each continuous geometric embedding to the closest codeword in the first-stage codebook, producing a coarse discrete embedding. We then compute the residual between the coarse embedding and the original continuous representation, and quantize it using the second-stage codebook to obtain a fine discrete embedding. The final discrete representation is formed by adding the coarse and fine components.

We further summarize the velocity estimation procedure during inference in \cref{alg:knn_velocity}. 
For each detection in the current frame, we first retrieve the $k$ nearest historical trajectory states in the spatial coordinate space. 
The velocity of the detection is then estimated by aggregating the historical velocities of these neighbors through a normalized weighted average. 
The weights combine a spatial proximity kernel and a directional consistency kernel, encouraging contributions from trajectories that are both spatially close and motion-aligned. 
The resulting velocity estimate is subsequently converted to the polar-coordinate representation and used to construct the geometric vector for quantization. 
This lightweight procedure provides a stable motion estimate for detections without explicit temporal correspondences while maintaining compatibility with the geometry quantization module.

\begin{algorithm}[t]
\caption{KNN-based Velocity Estimation for Inference}
\label{alg:knn_velocity}
\begin{algorithmic}[1]

\REQUIRE Detection center $\mathbf{c}_t^i=(c_{x,t}^i,c_{y,t}^i)$
\REQUIRE Historical trajectory states $\mathcal{H}=\{(\mathbf{c}_j,\mathbf{v}_j)\}$
\REQUIRE Neighbor number $k$, kernel scales $\sigma_s,\sigma_d$
\ENSURE Estimated velocity $\hat{\mathbf{v}}_t^i$

\STATE $\mathcal{N}_k \gets$ $k$ nearest states to $\mathbf{c}_t^i$ in spatial coordinate space

\FOR{each $j \in \mathcal{N}_k$}

    \STATE $d_{ij} \gets \|\mathbf{c}_t^i-\mathbf{c}_j\|_2^2$
           \hfill \textit{spatial distance}

    \STATE $\Delta\phi_{ij} \gets$ directional discrepancy between motion directions

    \STATE $s_{ij} \gets \exp(-d_{ij}/\sigma_s^2)$
           \hfill \textit{spatial kernel}

    \STATE $m_{ij} \gets \exp(\cos(\Delta\phi_{ij})/\sigma_d)$
           \hfill \textit{directional kernel}

    \STATE $\tilde w_{ij} \gets s_{ij} \cdot m_{ij}$

\ENDFOR

\STATE $w_{ij} \gets \tilde w_{ij} / \sum_{l \in \mathcal{N}_k} \tilde w_{il}$
       \hfill \textit{normalize weights}

\STATE $\hat{\mathbf{v}}_t^i \gets \sum_{j \in \mathcal{N}_k} w_{ij}\mathbf{v}_j$

\STATE \textbf{return} $\hat{\mathbf{v}}_t^i$

\end{algorithmic}
\end{algorithm}

\section{Visualization Results}
\label{sec:visualization}

\subsection{t-SNE Visualization}
\label{sec:t-SNE}
\begin{figure*}[t]
    \centering
    \begin{subfigure}[b]{1\textwidth}
        \centering
        \includegraphics[width=\linewidth]{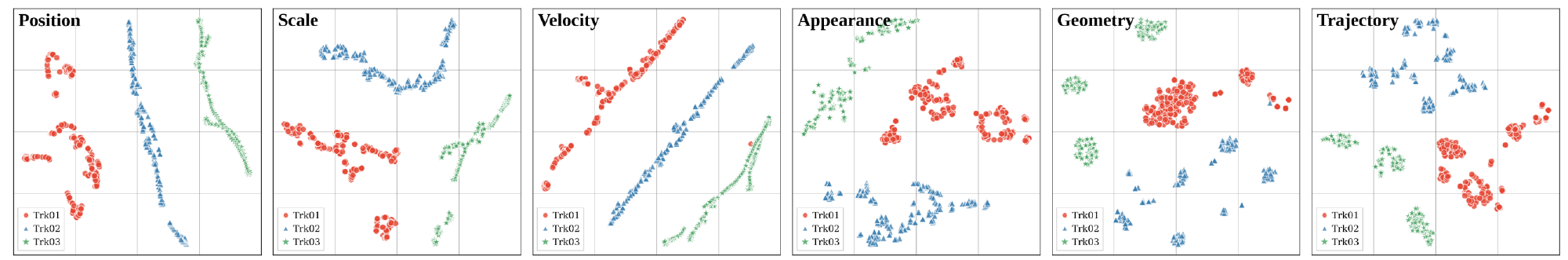}
        \caption{t-SNE~\cite{maaten2008visualizing} visualization of all frames in An3008 (BFT).}
    \end{subfigure}

    \begin{subfigure}[b]{1\textwidth}
        \centering
        \includegraphics[width=\linewidth]{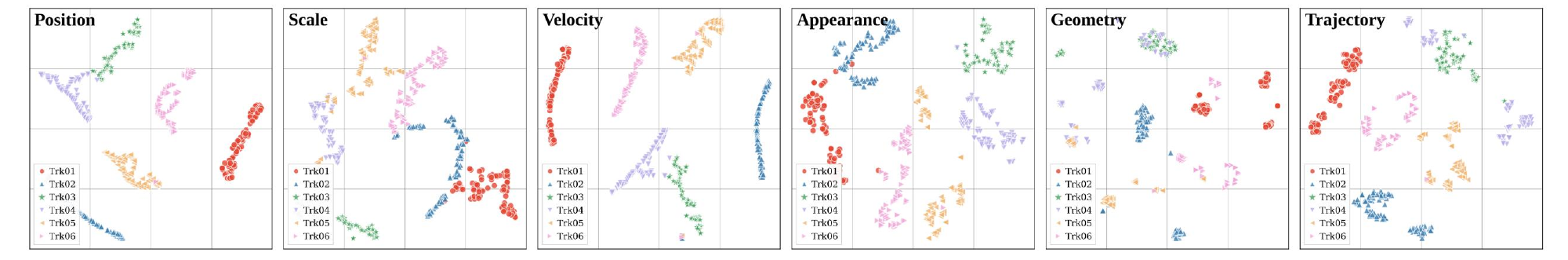}
        \caption{t-SNE~\cite{maaten2008visualizing} visualization of all frames in An1002 (BFT).}
    \end{subfigure}

    \begin{subfigure}[b]{1\textwidth}
        \centering
        \includegraphics[width=\linewidth]{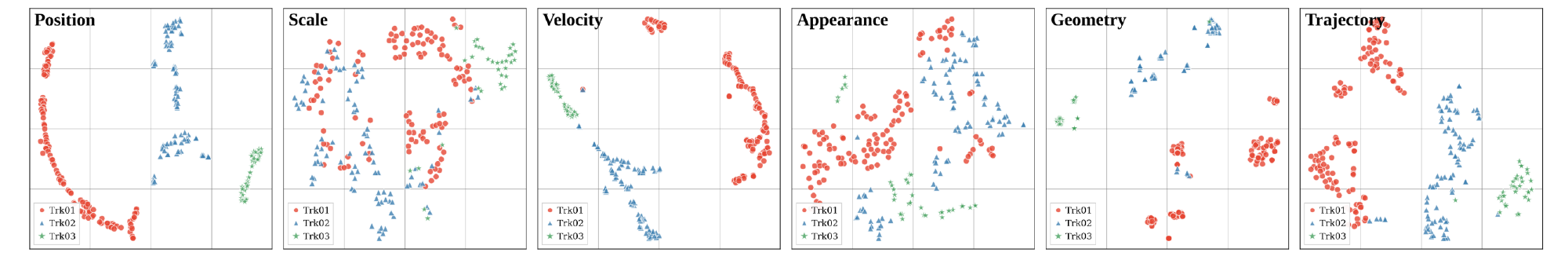}
        \caption{t-SNE~\cite{maaten2008visualizing} visualization of all frames in An6012 (BFT).}
    \end{subfigure}
    
    \caption{t-SNE~\cite{maaten2008visualizing} visualization of track embeddings across all frames in the BFT dataset. A unique combination of color and shape denotes each ID. We compare geometric embeddings, appearance embeddings, and their fused trajectory embeddings, highlighting the discriminative capacity of each representation.}
    \label{fig:tsne1}
\end{figure*}

Due to space limitations in the main text, we could not provide comprehensive t-SNE and tracking visualizations in the main paper. Here we present these visual results along with more detailed analyses. Fig.~\ref{fig:tsne1} shows the t-SNE embeddings for sequences An3008, An1005, An1002, and An6012 from the BFT dataset. For the geometric features, we observe that the position and velocity embeddings across video sequences generally follow spatial and motion consistency, remaining stable or changing gradually. This behavior provides reliable support for the subsequent association stage. In contrast, the size embeddings exhibit less clear separation, both in inter-class distance and intra-class compactness, suggesting potential directions for future improvement.

At a broader level, visual embeddings in animal-tracking scenarios often show weak structure, with inconsistent intra-ID and inter-ID distances. By comparison, incorporating our discrete geometric embeddings produces fused trajectory embeddings that display much clearer separation than appearance embeddings alone. This indicates that the proposed discrete geometric representation effectively complements visual cues in MAT tasks and contributes to more reliable association.

\subsection{Tracking Visualization}
\label{sec:tracking_visualization}

\begin{figure*}[t]
    \centering
    \begin{subfigure}[b]{1\textwidth}
        \centering
        \includegraphics[width=\linewidth]{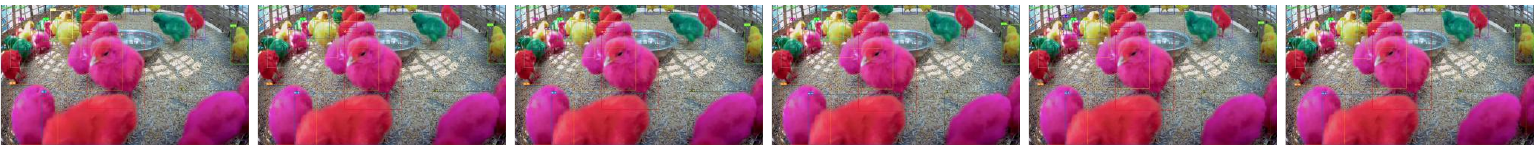}
        \caption{Chicken.}
    \end{subfigure}

    \begin{subfigure}[b]{1\textwidth}
        \centering
        \includegraphics[width=\linewidth]{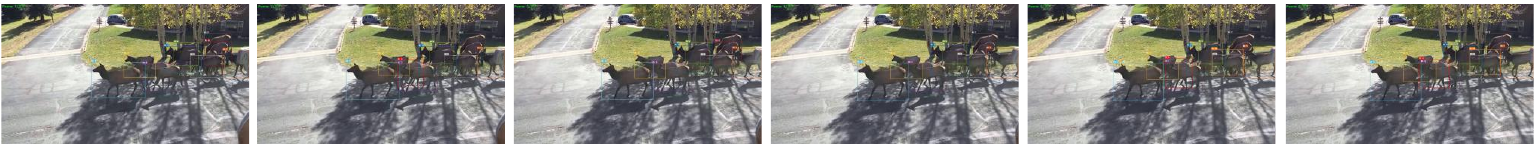}
        \caption{Deer.}
    \end{subfigure}

    \begin{subfigure}[b]{1\textwidth}
        \centering
        \includegraphics[width=\linewidth]{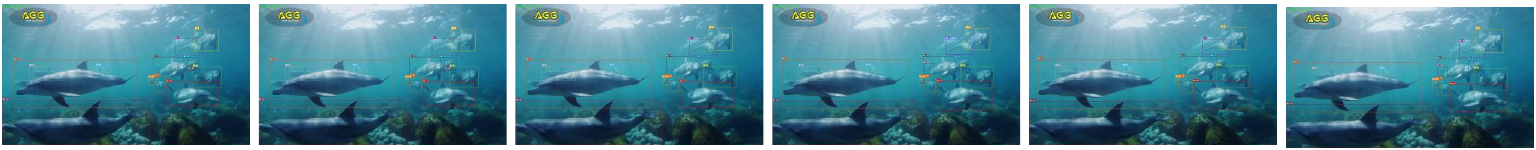}
        \caption{Dolphin.}
    \end{subfigure}

    \begin{subfigure}[b]{1\textwidth}
        \centering
        \includegraphics[width=\linewidth]{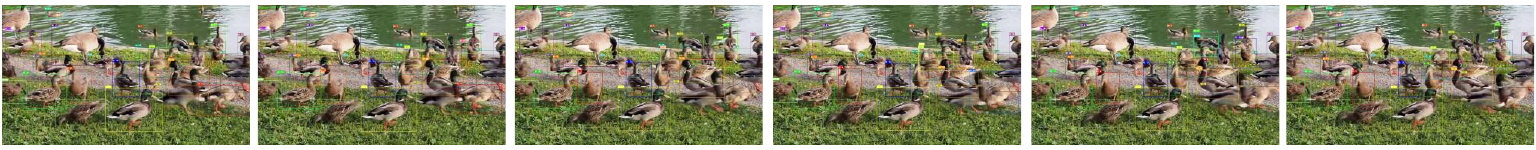}
        \caption{Duck.}
    \end{subfigure}

    \begin{subfigure}[b]{1\textwidth}
        \centering
        \includegraphics[width=\linewidth]{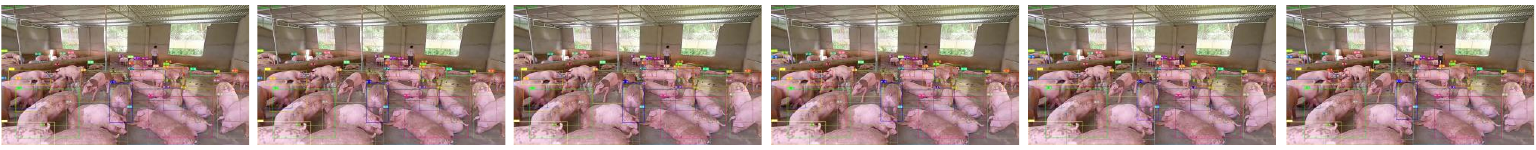}
        \caption{Pig.}
    \end{subfigure}

    \begin{subfigure}[b]{1\textwidth}
        \centering
        \includegraphics[width=\linewidth]{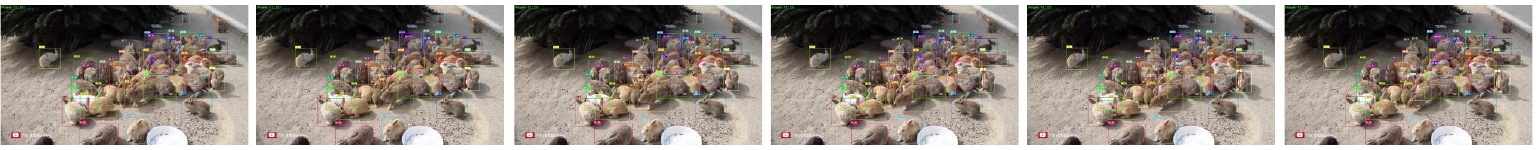}
        \caption{Rabbit.}
    \end{subfigure}
    
    \caption{Tracking visualization of HieDG on the AnimalTrack dataset.}
    \label{fig:track_visualization}
\end{figure*}

In this section, we also present six consecutive-frame tracking visualizations from the AnimalTrack dataset. Representative scenes, including chicken, deer, and dolphins, are shown in Fig.~\ref{fig:track_visualization}. We observe that tracking performance remains stable and reliable under natural conditions. However, in scenes such as chicken and pig, missed detections occur when animals appear near the image boundary or undergo partial occlusion, which prevents correct association in subsequent frames. In addition, under severe occlusion or significant appearance changes, such as ducks dipping their heads underwater or deer undergoing pronounced posture variation, HieDG may treat the target as a newly appearing individual and assign a new identity.

\section{Discussions}
\label{sec:Discussion}

Our HieDG framework provides an end-to-end paradigm that handles both object detection in animal scenes and identity association for tracking. Because appearance embeddings often show limited discrimination in animal scenarios, we integrate geometric embeddings into standard query-based tracking pipelines to strengthen the association stage. To cope with the substantial fluctuations in position, size, and velocity that are common in animal motion, we adopt a hierarchical residual quantization scheme that discretizes these geometric cues. The resulting geometric representation complements the visual embeddings, allowing one modality to support the other when its signal is weak or ambiguous.

At this stage, our geometric embedding design combines position, size, and velocity through linear mappings followed by concatenation. Future work may explore more expressive fusion strategies, including element-wise gated fusion, FiLM-style modulation, attention-based fusion such as cross-attention or self-attention, and graph-based message passing. These approaches could more effectively inject geometric information into end-to-end query-based tracking. In addition, the t-SNE visualizations in Appendix~\ref{sec:t-SNE} show that position and velocity form the main discriminative structure in our discrete geometric embeddings, while scale features remain less distinct. This suggests that more advanced size modeling, such as area–shape decomposition, log-space encoding, or ratio-based representations, could further improve performance.

Although our current study focuses on multi-animal tracking, we plan to extend the framework to broader tracking tasks. The proposed HieDG is not limited to animal scenes: its robust geometric modeling is also suitable for human and vehicle MOT. Since transformer-based tracking-by-query methods benefit from large-scale training data, we expect HieDG to perform even better as more comprehensive datasets become available.

%
%

%
%
\bibliographystyle{splncs04}
\bibliography{main}

\
\end{document}